\documentclass[11pt]{article}

\usepackage[final]{acl}

\usepackage{times}
\usepackage{latexsym}

\usepackage[T1]{fontenc}

\usepackage[utf8]{inputenc}

\usepackage{microtype}

\usepackage{graphicx}

\usepackage{amsmath}
\usepackage{amssymb}

\usepackage{comment}
\usepackage{booktabs} 
\usepackage{xspace}
\usepackage{multirow}
\usepackage{array}
\usepackage[table]{xcolor}
\usepackage{enumitem}
\usepackage{adjustbox}
\usepackage{subcaption}
\newcommand{\Linear}{\textbf{\textsc{\textcolor{black}{Linear}}}\xspace} 
\newcommand{\Lin}{\textbf{\textsc{\textcolor{black}{Lin}}}\xspace} 
\newcommand{\Hierarchical}{\textbf{\textsc{\textcolor{black}{Hierarchical}}}\xspace} 
\newcommand{\Hier}{\textbf{\textsc{\textcolor{black}{Hier}}}\xspace} 
\definecolor{mygreen}{rgb}{0.13, 0.55, 0.13}
\definecolor{myred}{rgb}{0.7, 0.1, 0.1}

\definecolor{ScoreLow}{HTML}{FFFFFF}
\definecolor{ScoreMid}{HTML}{FFFFFF}
\definecolor{ScoreHigh}{HTML}{FFFFFF}
\newcommand{\scorecell}[2]{%
  \ifdim #1 pt < 50pt
    \cellcolor{ScoreLow}{#1}%
  \else\ifdim #2 pt > 1pt
    \cellcolor{ScoreHigh}{#1}%
  \else
    \cellcolor{ScoreMid}{#1}%
  \fi\fi
}
\definecolor{fulloverlap}{HTML}{fe6100}
\newcommand{\fulloverlap}{\textcolor{black}{\textsc{Full Overlap}}\xspace}
\definecolor{1koverlap}{HTML}{ffb000}
\newcommand{\onekoverlap}{\textcolor{black}{\textsc{1k Overlap}}\xspace}
\definecolor{nooverlap}{HTML}{dc267f}
\newcommand{\nooverlap}{\textcolor{black}{\textsc{No Overlap}}\xspace}

\newcommand{\hidden}[1]{}
\newcommand{\update}[1]{#1}

\title{Once a Tree, Always a Tree? \\ Cross-lingual Transfer of Hierarchical Generalization in Language Models}
\author{
    Shin Yamaguchi \textsuperscript{$1$} \quad
    Masaya Taniguchi\textsuperscript{$2$} \\
    \textbf{Tatsuki Kuribayashi}\textsuperscript{$3,4$} \quad
    \textbf{Taro Watanabe}\textsuperscript{$1$} \quad
    \textbf{Yoshinobu Kano}\textsuperscript{$5$}
    \\[1ex]
    \textsuperscript{$1$}NAIST \quad
    \textsuperscript{$2$}RIKEN \quad
    \textsuperscript{$3$}MBZUAI \quad
    \textsuperscript{$4$}Tohoku University \quad
    \textsuperscript{$5$}Shizuoka University
    \\
     \texttt{yamaguchi.shin.yw4@naist.ac.jp} \quad
     \texttt{masaya.taniguchi@riken.jp}\\
     \texttt{tatsuki.kuribayashi@mbzuai.ac.ae}\quad
     \texttt{taro@is.naist.jp} \\
     \texttt{kano@inf.shizuoka.ac.jp}
}

\begin{document}
\maketitle
\begin{abstract}
Language models (LMs) often exhibit successful cross-lingual transfer through continued (pre)training, still their underlying process remains unclear.
One possibility is that LMs can leverage the evidence toward language-universal principles, namely, hierarchical structure, from a source language (L1) to a new language (L2), while many other non-linguistic explanations, e.g., improved initialization of position embeddings, can be potentially possible.
In this study, to quantify the cross-linguistic transferability of linguistic evidence in cross-lingual LM training, we introduce a new experimental design, namely, a cross-lingual poverty-of-stimulus (CPOS) setting that ablates how LMs' hierarchical generalization in L2 can be led by structural evidence provided in L1.
We contrast counterfactual L1 and L2 learning scenarios where the L2 corpus is ambiguous between linear and hierarchical grammar rules, while the amount of positive evidence for the hierarchical rule is controlled on the L1 side.
Our experimental results show that LMs can \update{partially}, yet not perfectly, transfer L1 structural evidence to L2\update{, yielding only a small but detectable hierarchical bias}, while the strength of transfer depends on language pairs.
We further ablate the conditions for the successful transferring, motivated by recent reports on the substantial impact of vocabulary overlap. 
We indeed find that vocabulary overlap is crucial even for the cross-lingual transfer of hierarchical syntactic generalization.
This suggests That this also explains the language-pair-dependent tendencies in the main experiments and suggests that the transfer of linguistic evidence in continued new-language learning is only conditionally achievable.
\end{abstract}

\section{Introduction}
\label{sec:introduction}
\begin{figure}[t]
    \centering
      \includegraphics[width=\linewidth]{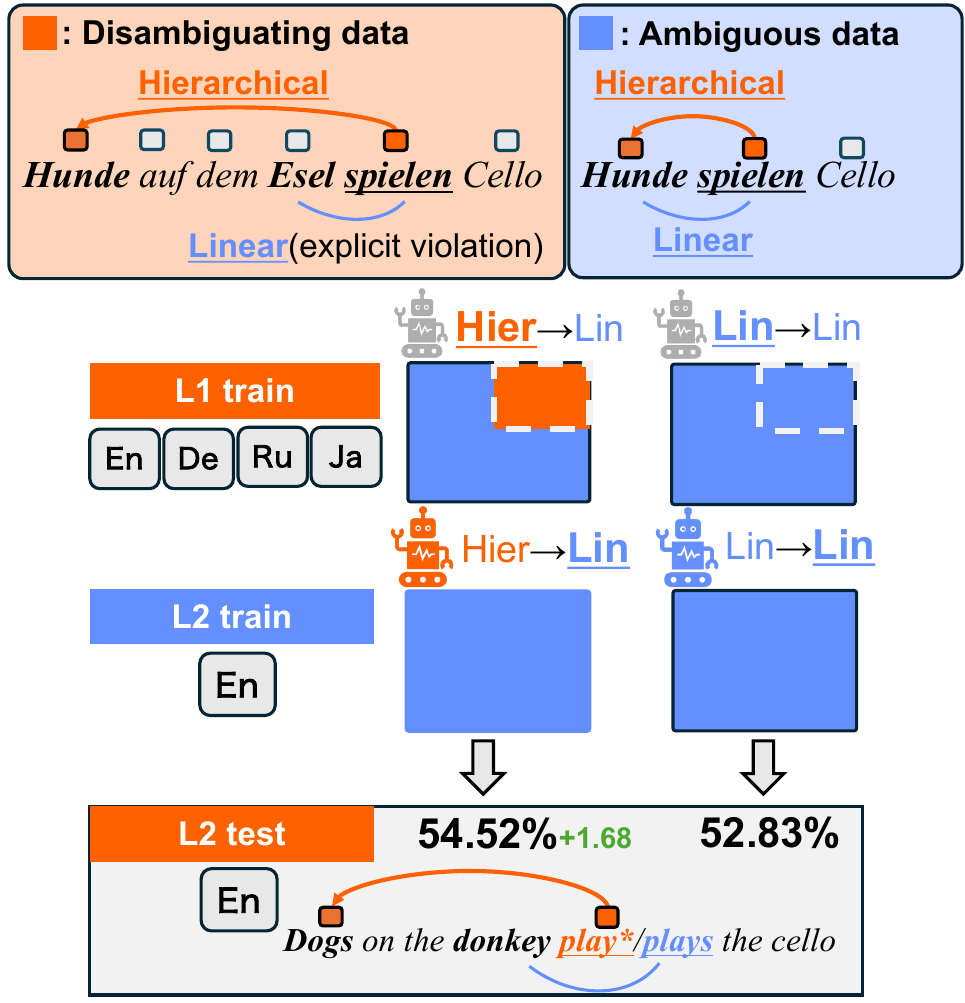}
      \caption{Overview of the experimental design. 
      A language model is first pretrained on an L1 corpus that contains positive evidence for the hierarchical rule \update{(\textbf{Disambiguating data})}, and then continued-pretrained on an L2 corpus from which such evidence has been removed \update{(\textbf{Ambiguous data})}. Any above-chance hierarchical-syntax performance at L2 must therefore reflect cross-lingual transfer of the rule acquired during L1 training.}
      \setlength{\belowcaptionskip}{-100pt}
      \label{fig:figure1}
\end{figure}
Cross-lingual transfer of language models (LMs) generally succeeds in learning a new language and, in practice, helps propagate LMs developed in one language to others~\cite{conneau-etal-2020-emerging}.
Even from a scientific perspective, cross-lingual transfer has served as a window into whether and how LMs encode language-universal properties from source data (L1) and use them for learning a new language (L2)~\cite{papadimitriou-jurafsky-2020-learning, ri-tsuruoka-2022-pretraining}.
This even potentially bridges LMs into psycholinguistics as a tool for characterizing human L2 learning, regarding LMs' cross-lingual transfer as a reference~\cite{oba-etal-2023-second, yadavalli-etal-2023-slabert, aoyama-schneider-2024-modeling}.
Despite its ubiquity, the underlying mechanisms and biases behind LMs' cross-lingual transfer remain unclear.
One prominent view is that LMs identify and abstract language-universal properties from evidence provided in L1 data, especially the hierarchical organization of natural-language structure, and then leverage this evidence in L2 learning.
Many other possibilities, however, can coexist alongside successful cross-lingual transfer, such as a mere scaling effect from increased pre-training data size or non-linguistic initialization artifacts, such as improved position embeddings learned from L1 data.
Studying a causal effect from L1 to L2 transfer learning requires controlled, counterfactual ablations.

In this paper, we conduct targeted ablations of LMs' L1 and L2 transfer learning scenarios to isolate the transfer of language-universal, hierarchical properties that have long been studied in linguistics~\cite{chomsky1980rules, legate2002empirical, linzen-etal-2016-assessing, gulordava-etal-2018-colorless}.
We specifically introduce a new experimental design, namely, a cross-lingual poverty-of-stimulus~(CPOS) setting, in which the L2 corpus is ambiguous between linear and hierarchical rules of L2 grammar, and positive evidence for the hierarchical rule is available only on the L1 side.
If LMs properly abstract the evidence in L1 to prefer hierarchical rules, increasing the amount of positive evidence on the L1 side will clearly improve the hierarchical generalization of L2 grammar.
We implement such experiments using subject-verb number agreement constructions (see Figure~\ref{fig:figure1}) and control the number of direct evidence disambiguating the hierarchical agreement rule from a position-based one, as well as indirect evidence from deeply nested constructions implicitly clarifying the hierarchical structure\update{\footnote{Code : \url{https://github.com/shin0729/cpos}; Data: \url{https://huggingface.co/datasets/shin0729/cpos}}}.

We perform experiments with four L1 languages~(i.e., English, German, Russian, and Japanese) to analyze language-dependent factors and evaluate LMs' generalization in the English language (L2).
Our experiments demonstrate that exposure to positive evidence in L1 indeed enhances, though not perfectly, the hierarchical generalization in L2, confirming the cross-lingual transfer of positive evidence.
We also observe some language-pair-dependent tendencies; for example, the German language as L1 leads to more successful generalization in L2 English than Japanese as L1.
To clarify these results, we further analyze the lexical overlap effect, which has been reported as a dominant factor in predicting the success of cross-lingual transfer in downstream tasks~\cite{kallini-etal-2025-false}.
Our analysis echoes the reported trends: eliminating vocabulary overlap, even using the same English corpora for both L1 and L2 but encoded with disjoint token IDs, leads to a clear failure of transfer.
This suggests that the transfer of positive linguistic evidence toward the language-universal, hierarchical structure is only conditionally successful, and the primary condition is vocabulary overlap.
That is, the vocabulary overlap is a fundamental path for successful cross-lingual transfer, even for the primitive language-generalization problem of syntactic rules, let alone downstream tasks~\cite{kallini-etal-2025-false}.

\section{Backgrounds}
\label{sec:background}
\subsection{Cross-lingual Transfer and the L2-LM Paradigm}
\label{subsec:cross_lingual_rw}
Existing works examine cross-lingual transfer both as a practical recipe for extending LMs to new languages and as a scientific probe into the kinds of transferable knowledge LMs acquire from L1 data and how this biases their L2 learning and processing. 
This L2-LM paradigm~\cite{aoyama-schneider-2024-modeling,oba-etal-2023-second,arnett-etal-2025-acquisition,yadavalli-etal-2023-slabert,gao-etal-2025-bliss} organizes the latter perspective, casting continued pretraining from L1 to L2 as a bilingual setting: varying L1 while holding L2 fixed isolates the effect of the source language on L2 outcomes, and the resulting L2 trajectories provide a baseline for comparison against human L2 acquisition. We adopt this design and combine it with the linguistic and non-linguistic factors discussed below in a poverty-of-stimulus setting targeting hierarchical generalization (\S\ref{subsec:hierarchical_generalization}).
Some studies show that typological distance between languages is related to the transfer effect~\cite{yadavalli-etal-2023-slabert,oba-etal-2023-second}. To probe this transferability, prior work uses Test for Inductive Bias via Language Model Transfer (TILT)-based methods that test whether inductive biases acquired during source-language training transfer to target-language evaluations~\cite{papadimitriou-jurafsky-2020-learning,aoyama-schneider-2024-modeling}, as well as crosslinguistic priming effects across language pairs~\cite{arnett-etal-2025-acquisition}.

A related line of work explicitly introduces inductive biases into LMs by pretraining them on non-natural inputs such as music or artificial languages~\cite{papadimitriou-jurafsky-2020-learning,ri-tsuruoka-2022-pretraining}, allowing controlled tests of whether source-language training induces a form of structural meta-learning for the target language.
Other studies examine non-linguistic factors, including different conditions of language exposure~\cite{constantinescu2025investigating,arnett-etal-2025-acquisition} and parameter sharing across languages~\cite{conneau-etal-2020-emerging}.

\paragraph{Vocabulary Overlap in Cross-Lingual Transfer}
\label{subsec:vocab_overlap_rw}
In addition to these factors, it has been shown that cross-lingual transfer in LMs depends on the degree of subword overlap between the paired-language vocabularies~\cite{kallini-etal-2025-false,skrjanec2026crosslingualtransferbilingualmodels}.
In particular, \citet{kallini-etal-2025-false} devise a controlled intervention in which the token IDs of the \emph{native overlap} are selectively offset to enforce disjoint L1 and L2 vocabularies, and show that the presence of shared vocabulary drives cross-lingual transfer, improving downstream task performance.
\citet{skrjanec2026crosslingualtransferbilingualmodels} likewise find that the cross-lingual alignment of embedding
spaces is largely driven by vocabulary sharing.  We adopt the same
form of overlap control (\S\ref{sec:analysis}).

\subsection{Hierarchical Generalization and Poverty-of-Stimulus Design}
\label{subsec:hierarchical_generalization}
\begin{figure*}[t]
    \centering
      \includegraphics[width=0.8\linewidth]{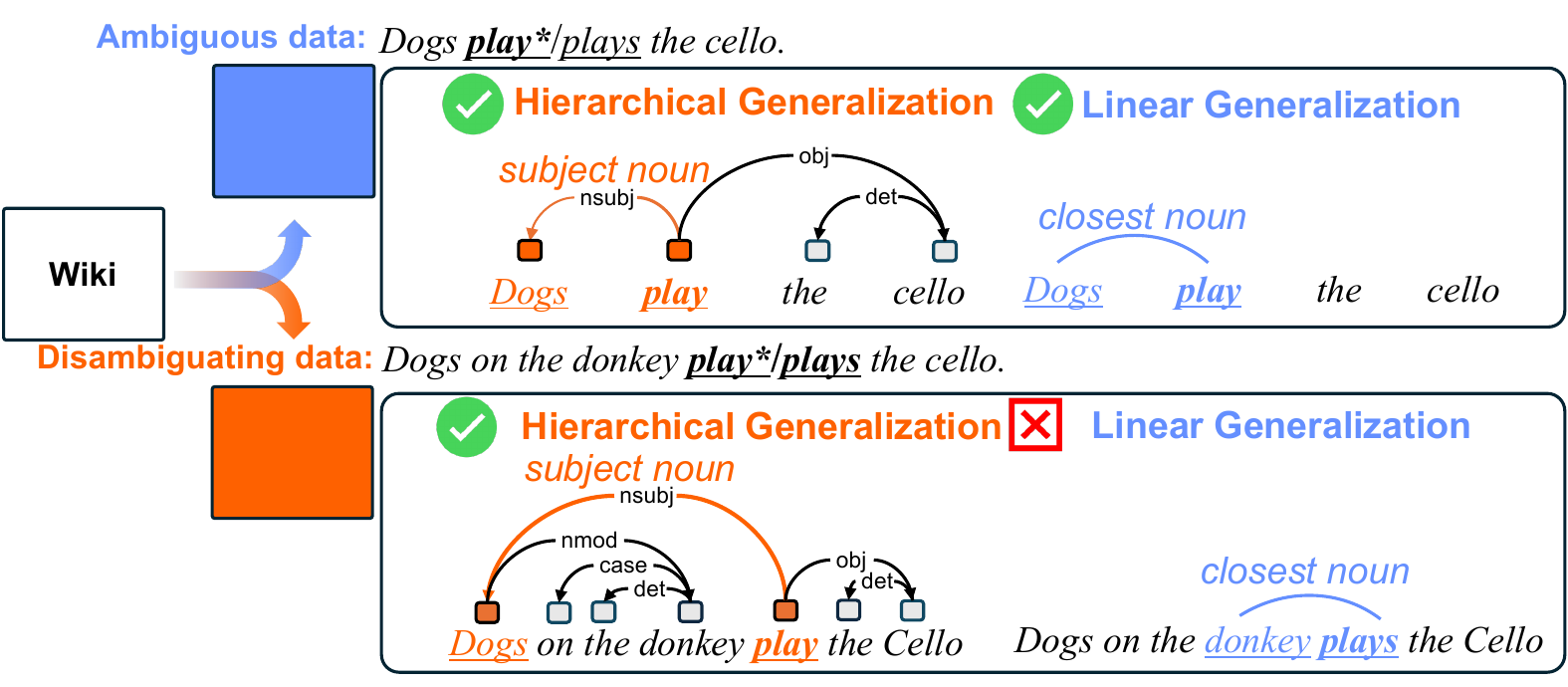}
      \caption{Example of \Hierarchical{} vs.\ \Linear{}
      generalization for subject-verb number agreement. With
      an attractor (\emph{donkey}) intervening between the
      plural subject (\emph{Dogs}) and the verb,
      \Linear{} predicts agreement with the linearly nearest
      noun (\emph{plays}), whereas \Hierarchical{} predicts
      agreement with the syntactic subject (\emph{play}).}
      \setlength{\belowcaptionskip}{-100pt}
      \label{fig:hg_exsample}
\end{figure*}
Hierarchical generalization of language learners has typically been debated and studied within the
\emph{poverty of stimulus} (POS)
paradigm~\cite{mccoy-etal-2020-syntax,mueller-etal-2022-coloring,ahuja-etal-2025-learning,mulligan-etal-2021-structure,kuribayashi-baldwin-2025-vision}:
training data are curated intentionally ambiguous between a linear rule~(\Linear) and a hierarchical rule~(\Hierarchical), and the
model is then evaluated on held-out constructions where the
two rules make divergent predictions, revealing which rule the
model has adopted. 
When a number-mismatched noun---an
\emph{attractor}---intervenes between a plural subject and its
verb (e.g., \emph{Dogs on the donkey play/plays the cello};
Figure~\ref{fig:hg_exsample}), \Linear{} predicts agreement
with the linearly nearest noun, whereas \Hierarchical{}
predicts agreement with the syntactic subject. Attractor-bearing
sentences thus serve as direct positive evidence of hierarchical
rule~\cite{linzen-etal-2016-assessing,gulordava-etal-2018-colorless, kuribayashi-baldwin-2025-vision}.

A growing body of work has identified factors that drive
hierarchical generalization in LMs, including the amount of
positive evidence in the training
data~\cite{mccoy-etal-2020-syntax,mueller-linzen-2023-plant,oba-etal-2024-language,misra-mahowald-2024-language,guven-etal-2025-syntactic,kuribayashi-baldwin-2025-vision},
the depth of clausal embedding~\cite{qin-etal-2025-data}, the
choice of training objective~\cite{ahuja-etal-2025-learning},
training duration beyond the point of
overfitting~\cite{murty-etal-2023-grokking}, and access to
additional modalities such as
vision~\cite{kuribayashi-baldwin-2025-vision}. By contrast,
LMs trained on small, developmentally plausible inputs such
as child-directed speech have been reported to fail to
acquire the hierarchical
rule~\cite{yedetore-etal-2023-poor,padovani-etal-2025-child}. All of these studies,
however, share a monolingual design in which the language
that provides positive evidence and the language on which
generalization is evaluated coincide, leaving open whether the
acquired hierarchical bias is tied to a particular language or
instead reflects a language-general structural abstraction.

\section{Tasks and Data}
\label{sec:task_data}

\subsection{Cross-lingual Poverty-of-Stimulus (CPOS) Setting}
The \update{POS} settings have typically been employed to analyze the inductive biases of LMs~\cite{misra-mahowald-2024-language, oba-etal-2024-language,yang2026unifiedassessmentpovertystimulus}.
The original POS argument in language acquisition in human infants claims the impossibility of identifying the correct linguistic rules underlying (infinite) linguistic input from finite observation, let alone limited exposure in human infants, necessitating strong inductive biases in human learners~\cite{legate2002empirical, crain1987structure}.
Motivated by this argument, to assess the LMs' inductive biases, LMs are trained on data that are generally compatible with several generalization rules.\footnote{Our study borrows the experimental settings of the POS paradigm and does not claim implications for the original POS debate.}
Controls on the availability of evidence during training and their evaluation clarify, for example, which rules LMs' inductive biases prefer and how sensitive LMs are to such evidence when inducing rules from data.

We extend this setting to be cross-lingual; positive evidence for a specific linguistic rule governing L1 and L2 grammars is only available in the L1 corpus.
This clarifies the cross-lingual transfer ability of LMs by examining whether they can leverage L1 evidence for L2 generalization by controlling the number of instances with such evidence.
Specifically, we first pretrain LMs on the L1 corpus with positive evidence, and continuously train them on the L2 corpus without such evidence, and evaluate the generalization on L2.
We change the number of instances with certain positive evidence on the L1 side, and assess how sensitive the L2 generalization is to that L1 factor.

\subsection{Evaluation Task (L2)}
\label{subsec:eval_task}
Following prior works~\cite{linzen-etal-2016-assessing,gulordava-etal-2018-colorless,jumelet-etal-2026-multiblimp,kuribayashi-baldwin-2025-vision}, we select the subject-verb agreement judgment as a testbed on the L2 side to evaluate the hierarchical generalization of LMs after cross-lingual transfer.
The motivation behind this selection is that, first, as shown in Figure~\ref{fig:figure1}, one can clearly define instances (positive evidence) that disambiguate the hierarchical rule and the other ones that do not.
Simply put, if the subject noun phrase (NP) is the closest NP to the verb, there are two possible generalizations: (i) the grammatical number of the verb should match that of its grammatical subject (\Hierarchical{} generalization), or (ii) it should match the closest NP to the verb, based on linear position (\Linear{} generalization).
The L1 training data should be controlled with respect to how informative the data are to disambiguate the \Hierarchical{} and \Linear{} rules.
For the evaluation, one can distinguish whether the model generalizes the agreement linearly or hierarchically, using examples with \update{an attractor} (\update{e.g., \textit{donkey}} in Figure~\ref{fig:figure1}).
Second, the subject-verb agreement phenomenon is widely attested in many languages, which enables scaling the experiments in multilingual settings, similarly to MultiBLiMP~\cite{jumelet-etal-2026-multiblimp}.

\subsection{Positive Evidence for Hierarchical Generalization (L1)}\label{sec:hierarchy_definition}
We then define positive (direct and indirect) evidence for the hierarchical generalization of subject-verb agreement with two criteria.
In the ablation experiments, the instances matching the following criteria are removed (or not) from the L1 corpus.

\paragraph{Attractors.} 
Following
\citet{linzen-etal-2016-assessing} and
\citet{jumelet-etal-2026-multiblimp}, if noun phrases (NPs) are embedded between the subject NP and its verb (attractors), and the grammatical number of the subject NP and attractor NPs differ, such an instance clarifies the agreement rule targeted in Section~\ref{subsec:eval_task}, serving as positive evidence.
Note that there are several agreement phenomena similar to grammatical number, for example, grammatical gender agreement, which can also potentially disambiguate the hierarchical rules between NP and verb.
It is difficult to enumerate and identify all features in different L1 language corpora, and we adopt a somewhat conservative
criterion to remove all intervening NPs between NP and verb, regardless of their morphological features.
Still, related constructions such as pre-/post-positional attachment can appear in many other constructions, and thus LMs are required to perform compositional generalization and identify hierarchical generalization rules.

\paragraph{Embedding depth.} 
Another possible source of indirect evidence for the hierarchical generalization is not at the word--word level with morphological features, but also at the clause level; for example, center-embedding clarifies the hierarchical organization of sentence structure.
To maximize the difference in the availability of cues for hierarchical generalization, we also remove such instances from the L1 corpus by counting the embedding depth of each sentence and filtering out the ones with an embedding depth of more than N.

\subsection{L1 Training Data Creation}
\label{subsec:data_construction}
We create the L1 training data for four different, typologically diverse languages (i.e., English, German, Russian, and Japanese), and use Wikipedia\footnote{\url{https://huggingface.co/datasets/wikimedia/wikipedia}} as a source of the corpora.
We convert a base corpus into two versions of training data for each language, which differ only in the availability of positive evidence: a \update{POS} variant~(\textbf{Ambiguous data}) and an evidence-enriched counterpart~(\textbf{Disambiguating data}). 
Specifically, we first parse the sentences in the base corpus with \update{Stanza}~\cite{qi-etal-2020-stanza} and classify each sentence into either a \emph{positive-evidence sentence} that supports hierarchical generalization or a \emph{non-evidential sentence} that supports both hierarchical and linear generalizations \update{(Figure~\ref{fig:hg_exsample})}, according to the two criteria explained in Section~\ref{sec:hierarchy_definition}, resulting in the following two versions of training data
\footnote{\update{Detailed statistics of the resulting training corpora are provided in Appendix~\ref{app:corpus_stats}.}} with the same \update{token counts}: 
\begin{description}
  \item[\update{\textbf{Ambiguous data}}:] the base corpus without all positive-evidence sentences, where a learner is exposed only to sentences that basically lack evidence for hierarchical generalization.
  \item[\update{\textbf{Disambiguating data}}:] a corpus with positive evidence by replacing sentences in \textbf{Ambiguous data} with positive-evidence sentences.
\end{description}
We train different LMs by changing only their training data (Ambiguous data or Disambiguating data) and analyze the differences in their generalization patterns, which are attributed to the presence of positive evidence in L1. 
The implementation details of the parsing and classification are shown in the Appendix~\ref{app:corpus_construction_details}.

\section{Experimental Settings}
\label{sec:method}
We test the cross-lingual transfer of hierarchical generalization by manipulating two factors of L1 training: the level of exposure to \textbf{Disambiguating data}, and the choice of L1.

\subsection{Condition 1: Exposure to Positive Evidence in L1} 
To assess whether L1 exposure to positive evidence drives transfer, we contrast two L1 training regimes that differ only in the availability of positive evidence during L1 training, adopting the following two scenarios:
\begin{description}
  \item [\Hier $\to$ \Lin:] L1 training data are sampled from \textbf{Disambiguating data}, and L2 training data from \textbf{Ambiguous data}.
  \item [\Lin $\to$ \Lin:] both L1 and L2 training data are sampled from \textbf{Ambiguous data}; this serves as the baseline against which the \Hier $\to$ \Lin condition is compared.
\end{description}

\subsection{Condition 2: Choice of L1}
To assess whether transfer is modulated by typological properties of the source language, we vary L1 while holding L2 fixed at English. 
We select the following four languages that span a range of typological properties relevant to subject-verb agreement and the licensing of attractor configurations as L1:
\begin{description}
\item[L1 English] Given that L2 is also English, this condition serves as a within-language reference for transfer. 
\item[L1 German] German is a V2 language that, like English, licenses subject-verb agreement configurations involving attractors.
\item[L1 Russian] Russian also licenses attractor-bearing subject-verb agreement, but its considerably freer word order may weaken the attractor-as-evidence signal relative to English or German.
\item[L1 Japanese] Japanese exhibits very few attractor-bearing subject-verb agreement constructions.
\end{description}
These two conditions result in eight conditions in a $\{\Hier \to \Lin,\,\Lin\to\Lin\}\times\{\mathrm{En,\,De,\,Ja,\,Ru}\}$ design.

\subsection{Evaluation Metrics}
\label{subsec:evaluation_metrics}
\paragraph{MPP benchmarks.} \update{We evaluate LMs, following the minimal-pair paradigm (MPP), and use accuracy as our primary evaluation metric.}
For each minimal pair $(s^{+}, s^{-})$ in a benchmark $\mathcal{B}$, where $s^{+}$ is the grammatical sentence and $s^{-}$ its minimally differing ungrammatical counterpart, we count the item as correct when the model assigns lower sentence-level perplexity to $s^{+}$ than to $s^{-}$. 
Accuracy is the proportion of correct preferences over $\mathcal{B}$:
\begin{equation}
  \mathrm{acc} = \frac{1}{|\mathcal{B}|}
  \sum_{(s^{+}, s^{-})\, \in\, \mathcal{B}}
  \mathbf{1}\!\left[\mathrm{PPL}(s^{+}) < \mathrm{PPL}(s^{-})\right].
  \label{eq:acc}
\end{equation}
We evaluate on two complementary benchmarks:
MultiBLiMP~\citep{jumelet-etal-2026-multiblimp}, constructed from naturalistic multilingual corpora, and CLAMS~\citep{mueller-etal-2020-cross}, a multilingual benchmark of synthetic minimal pairs.
While we only use the English part of these benchmarks, we selected them as they target the subject-verb agreement phenomena.
We denote the value of Equation~\eqref{eq:acc} obtained under training condition $c$ as $\mathrm{acc}_{c}$.
CLAMS provides a large, systematically controlled set of synthetic minimal pairs targeting attractor configurations, whereas MultiBLiMP comprises far fewer naturally occurring minimal pairs whose lexical frequency and collocational confounds inflate the variance of accuracy estimates. 
We therefore treat CLAMS as our primary benchmark and use MultiBLiMP as a directional check.
We report, in addition to per-condition accuracy, the difference between the \Hier$\to$\Lin and \Lin$\to$\Lin conditions for each L1:
\begin{equation}
  \Delta\mathrm{acc} = \mathrm{acc}_{\Hier\to\Lin}
  - \mathrm{acc}_{\Lin\to\Lin}\;\;\mathrm{,}
  \label{eq:delta-acc}
\end{equation}
which isolates the contribution of L1 exposure to positive evidence; a positive $\Delta\mathrm{acc}$ is interpreted as evidence that the hierarchical generalization acquired in L1 has transferred to L2.
\subsection{Detailed Setup}
\label{subsec:setup}
\paragraph{Training Data.} Following the experimental design described in Section~\ref{sec:task_data}, we construct the L1 training data as a 50:50 mixture of \textbf{Disambiguating} and \textbf{Ambiguous} sentences\update{, while sampling the L2 training data} exclusively from the \textbf{Ambiguous} ones. 
In both L1 and L2 corpus, training and validation sets contain 50M and 5M tokens, respectively.

\paragraph{Models.} 
All models adopt the GPT-2-small architecture~\citep{radford2019language}, a standard Transformer language model trained with an autoregressive objective. 
Each model is trained from scratch on the L1 corpus for 8 epochs using this objective. After freezing\update{\footnote{We report results under the corresponding unfrozen setting in Appendix~\ref{app:unfreezing}.}}the transformer layers, pretraining continues on the L2 corpus for an additional 8 epochs, starting from the same checkpoint. Because each stage uses a 50M-token corpus, the model is exposed to a cumulative total of $50\text{M}\times8 = 400\text{M}$ tokens per stage. 
Full hyperparameter settings are reported in Appendix~\ref{app:hyperparameters}.

We train three models in each setting with different random seeds, and the results are averaged across these three runs.

\paragraph{Tokenizer.} \label{subsec:tokenizer}We train a single byte-level BPE tokenizer on the concatenation of the L1 and L2 corpora, with a vocabulary size of 64{,}000.

\section{Experimental Results}
\label{sec:experiment_and_result}
\begin{figure*}[t]
    \centering
      \includegraphics[width=\linewidth]{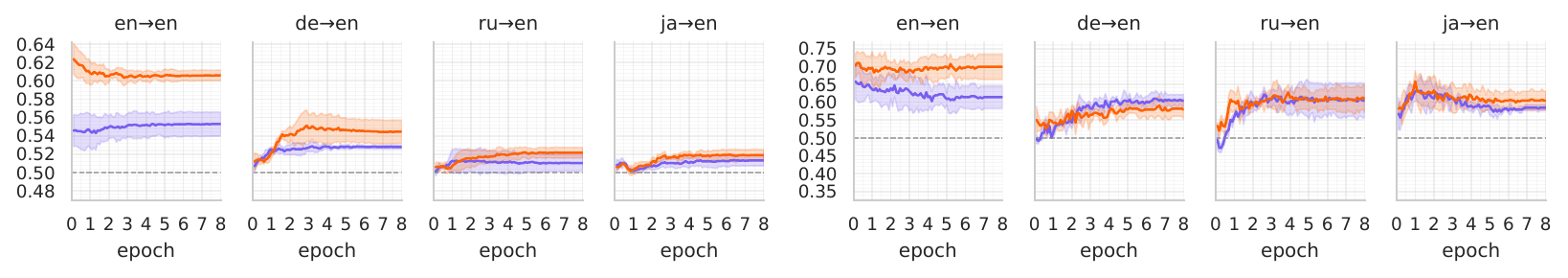}
      \caption{Learning trajectories on CLAMS (Left) and MultiBLiMP (Right) over L2-training epochs under the \fulloverlap{} condition. Orange lines: $\Hier\to\Lin$; blue lines: $\Lin\to\Lin$ baseline.
      Lines show the seed mean; shaded bands indicate $\pm 1$ SD across three random seeds.
      The chance rate is $0.50$.}
      \setlength{\belowcaptionskip}{-100pt}
      \label{fig:figure_acc_clams}
\end{figure*}

\begin{table}[t]
\centering
\small
\begin{tabular}{c|cc}
\toprule
L1 & \Lin $\to$ \Lin & \Hier $\to$ \Lin \\
\midrule
\multicolumn{3}{c}{\textbf{CLAMS}} \\
\midrule
\textit{En$\rightarrow$En} & \scorecell{56.04}{0.00} & \scorecell{61.49}{5.45} ($\Delta$ \textcolor{mygreen}{+5.45}) \\
\textit{De$\rightarrow$En} & \scorecell{52.83}{0.00} & \scorecell{54.52}{1.68} ($\Delta$ \textcolor{mygreen}{+1.68}) \\
\textit{Ru$\rightarrow$En} & \scorecell{51.07}{0.00} & \scorecell{52.11}{1.04} ($\Delta$ \textcolor{mygreen}{+1.04}) \\
\textit{Ja$\rightarrow$En} & \scorecell{51.24}{0.00} & \scorecell{51.77}{0.53} ($\Delta$ \textcolor{mygreen}{+0.53}) \\
\midrule
\multicolumn{3}{c}{\textbf{MultiBLiMP}} \\
\midrule
\textit{En$\rightarrow$En} & \scorecell{61.39}{0.00} & \scorecell{69.92}{8.53} ($\Delta$ \textcolor{mygreen}{+8.53}) \\
\textit{De$\rightarrow$En} & \scorecell{60.45}{0.00} & \scorecell{57.91}{-2.54} ($\Delta$ \textcolor{myred}{-2.54}) \\
\textit{Ru$\rightarrow$En} & \scorecell{60.45}{0.00} & \scorecell{61.02}{0.56} ($\Delta$ \textcolor{mygreen}{+0.56}) \\
\textit{Ja$\rightarrow$En} & \scorecell{58.47}{0.00} & \scorecell{60.45}{1.98} ($\Delta$ \textcolor{mygreen}{+1.98}) \\
\bottomrule
\end{tabular}
\caption{Accuracy (seed-mean of the final checkpoint, scored over tokens seen in L1 and/or L2 training data (OOV excluded)) under full vocabulary overlap, for benchmarks CLAMS, MultiBLiMP. The $\Hier\to\Lin$ column appends $\Delta=(\Hier\to\Lin)-(\Lin\to\Lin)$ at the same overlap (\textcolor{mygreen}{green}=positive, \textcolor{myred}{red}=negative).}
\label{tab:result_main_clams_multiblimp_seen}
\end{table}
We report the L2 (English) accuracy score in CLAMS and MultiBLiMP in Table~\ref{tab:result_main_clams_multiblimp_seen}; the corresponding learning trajectories through epochs are shown in \update{Figure}~\ref{fig:figure_acc_clams}.

We first confirm whether L1 exposure to positive evidence yields above-chance L2 hierarchical-generalization accuracy. 
Under $\Lin\to\Lin$, accuracy hovers around chance ($51.1$--$56.0\%$) for all four L1 languages on CLAMS, confirming that under our design, the model fails to perform hierarchical generalization from \textbf{Ambiguous data} alone.
By contrast, $\Hier\to\Lin$ yields $\Delta\mathrm{acc}>0$ for every L1\update{, most prominently for English ($+5.45$), followed by German ($+1.69$) and Russian ($+1.04$; Table~\ref{tab:result_main_clams_multiblimp_seen})}.
As neither positive evidence is at the L2 learning phase, these gains must be attributed to the positive evidence provided in L1: linguistic evidence toward hierarchical generalization \emph{does}, more or less, transfer across languages.
The magnitude of transfer varies with the L1 language.
\update{$\Delta\mathrm{acc}$ on CLAMS decreases monotonically from English ($+5.45$) $>$ German ($+1.68$) $>$ Russian ($+1.04$) $>$ Japanese ($+0.53$).} 
This gradient broadly aligns with the typological distance of each L1 from L2 English, which will be further clarified in  \S\ref{sec:analysis}.

On the MultiBLiMP dataset, English as L1 alone exhibits an even larger positive effect \update{than} on CLAMS (\update{$+8.53$}), reproducing the positive transfer evaluated on naturalistic data. 
The corresponding effects for German, Russian, and Japanese are $-2.54$, $+0.57$, and $+1.98$, respectively.
MultiBLiMP contains far fewer items than CLAMS, which may bring somewhat conservative results.
Notably, these transfer effects are not pronounced at any particular epoch in L2 training; rather, they become stable after several epochs (see \update{Figure}~\ref{fig:figure_acc_clams}).

\section{Analysis: Vocabulary Overlap}
\label{sec:analysis}

\update{The previous experiments establish that LMs can, to a small extent, leverage the positive evidence observed in L1 data for L2 learning.
However, these results alone do not establish that the observed differences arise specifically from the transfer of a hierarchical bias, as alternative explanations remain possible, including effects driven simply by lexical sharing between L1 and L2.
We therefore extend our experiments by explicitly controlling vocabulary overlap and conduct statistical analysis.}

\paragraph{Controlling Vocabulary Overlap}
\update{We hypothesize that shared vocabulary likewise facilitates the transfer of hierarchical generalization by providing cross-lingual anchors through which an L1-acquired syntactic rule can be reused in L2; removing overlap should therefore suppress transfer.}

\update{To test this hypothesis, following \citet{kallini-etal-2025-false}, let $V_{L_1}$ and $V_{L_2}$ denote the sets of tokenizer tokens that occur in the L1 and L2 corpora, respectively, and let $O=V_{L_1}\cap V_{L_2}$ denote their native overlap. We control the subset $O'\subseteq O$ that retains shared token IDs across L1 and L2.}
See Table~\ref{tab:vocab-overlap} for the resulting effective vocabulary sizes. 

We compare three conditions: (i) the L1 and L2 tokenizer vocabularies are fully shared ($O' = O$), corresponding to the experiments up to \S\ref{sec:experiment_and_result} ( \fulloverlap); (ii) only $|O'| = 1{,}000$ tokens of the native overlap are retained as shared, with the remainder offset, leaving only a minimal lexical bridge between the two languages (\onekoverlap); and (iii) all tokens in $O$ are offset ($O' = \emptyset$), so that L1 and L2 hold fully disjoint vocabulary sets (\nooverlap).
That is, for example, in the En$\rightarrow$En setting, the L2 English corpus is re-tokenized with a new tokenizer, and these IDs are appended to the L1 tokenizer as new IDs (e.g., \textit{the} in L1 and L2 have different IDs).
This is a somewhat extreme counterfactual baseline scenario; the learner learns a new language with perfectly the same syntactic rules as the L1, and only the vocabulary is different.
The implementation details of disjoint and control overlap are shown in the Appendix~\ref{app:details_of_overlap}.
\paragraph{Statistical Analysis}
\update{For each benchmark, we fit a Bayesian Bernoulli-logit mixed-effects model to item-level correctness across all 72 independently trained runs: \texttt{correct} $\sim$ \texttt{L1\_evidence} $*$ \texttt{overlap} $*$ \texttt{L1\_language} $+$ $(1\,|\,\texttt{run})$ $+$ $(1\,|\,\texttt{item})$; the full parameterization, priors, and convergence diagnostics are reported in Appendix~\ref{app:glmm_details}.}
\update{The model estimates $\Delta_{\mathrm{hier}}$ as the \Hier$\to$\Lin{} minus \Lin$\to$\Lin{} contrast within each overlap $\times$ L1-language combination; we report estimates (posterior medians) and equal-tailed 95\% credible intervals (CrIs) on the log-odds scale, where positive estimates indicate higher odds of a correct L2 hierarchical-syntax preference after exposure to positive evidence in L1 and an effect is described as credible when its CrI excludes zero.}

\begin{figure*}[t]
    \centering
      \includegraphics[width=\linewidth]{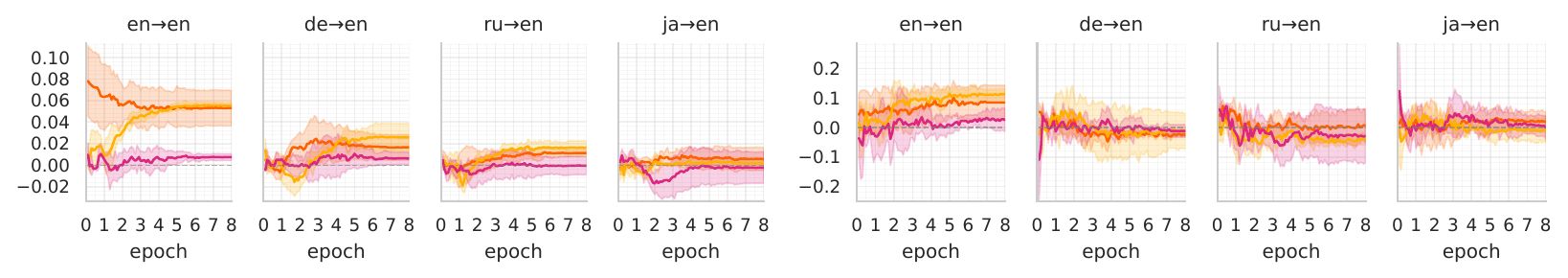}
      \caption{$\Delta\mathrm{acc}$ on CLAMS (Left) and MultiBLiMP (Right) broken down by L1, vocabulary-overlap condition, and L2-training epoch.
      Lines show the seed mean; shaded bands indicate $\pm 1$ SD across three random seeds.}
      \setlength{\belowcaptionskip}{-100pt}
      \label{fig:clams_lang_delta}
\end{figure*}

\paragraph{Results}
\begin{table*}[h]
\centering
\small
\setlength{\tabcolsep}{3pt}
\begin{tabular}{llrrrlll}
\toprule
& & \multicolumn{3}{c}{\Lin $\to$ \Lin} & \multicolumn{3}{c}{\Hier $\to$ \Lin} \\
\cmidrule(r){1-1} \cmidrule(lr){2-2} \cmidrule(lr){3-5} \cmidrule(lr){6-8}
Data & L1$\rightarrow$ L2 & \multicolumn{1}{c}{full} & \multicolumn{1}{c}{1K} & \multicolumn{1}{c}{no} & \multicolumn{1}{c}{full} & \multicolumn{1}{c}{1K} & \multicolumn{1}{c}{no} \\
\cmidrule(r){1-1} \cmidrule(lr){2-2} \cmidrule(lr){3-5} \cmidrule(lr){6-8}
\multirow{4}{*}{\textbf{CLAMS}} & \textit{En$\rightarrow$En} & \scorecell{56.04}{0.00} & \scorecell{55.61}{0.00} & \scorecell{51.61}{0.00} & \scorecell{61.49}{5.45} ($\Delta$ \textcolor{mygreen}{+5.45}) & \scorecell{61.37}{5.76} ($\Delta$ \textcolor{mygreen}{+5.76}) & \scorecell{52.39}{0.79} ($\Delta$ \textcolor{mygreen}{+0.79}) \\
 & \textit{De$\rightarrow$En} & \scorecell{52.83}{0.00} & \scorecell{52.78}{0.00} & \scorecell{51.42}{0.00} & \scorecell{54.52}{1.68} ($\Delta$ \textcolor{mygreen}{+1.68}) & \scorecell{55.46}{2.69} ($\Delta$ \textcolor{mygreen}{+2.69}) & \scorecell{52.19}{0.77} ($\Delta$ \textcolor{mygreen}{+0.77}) \phantom{} \\
 & \textit{Ru$\rightarrow$En} & \scorecell{51.07}{0.00} & \scorecell{51.39}{0.00} & \scorecell{50.92}{0.00} & \scorecell{52.11}{1.04} ($\Delta$ \textcolor{mygreen}{+1.04}) & \scorecell{53.03}{1.64} ($\Delta$ \textcolor{mygreen}{+1.64}) & \scorecell{51.07}{0.14} ($\Delta$ \textcolor{mygreen}{+0.14}) \phantom{***} \\
 & \textit{Ja$\rightarrow$En} & \scorecell{51.24}{0.00} & \scorecell{51.62}{0.00} & \scorecell{51.85}{0.00} & \scorecell{51.77}{0.53} ($\Delta$ \textcolor{mygreen}{+0.53}) & \scorecell{51.75}{0.13} ($\Delta$ \textcolor{mygreen}{+0.13}) \phantom{} & \scorecell{51.59}{-0.26} ($\Delta$ \textcolor{myred}{-0.26}) \phantom{*} \\
\cmidrule(r){1-1} \cmidrule(lr){2-2} \cmidrule(lr){3-5} \cmidrule(lr){6-8}
\multirow{4}{*}{\textbf{MultiBLiMP}} & \textit{En$\rightarrow$En} & \scorecell{61.39}{0.00} & \scorecell{61.11}{0.00} & \scorecell{51.94}{0.00} & \scorecell{69.92}{8.53} ($\Delta$ \textcolor{mygreen}{+8.53}) \phantom{***} & \scorecell{72.36}{11.25} ($\Delta$ \textcolor{mygreen}{+11.25}) \phantom{*} & \scorecell{54.47}{2.53} ($\Delta$ \textcolor{mygreen}{+2.53}) \phantom{***} \\
 & \textit{De$\rightarrow$En} & \scorecell{60.45}{0.00} & \scorecell{59.04}{0.00} & \scorecell{57.34}{0.00} & \scorecell{57.91}{-2.54} ($\Delta$ \textcolor{myred}{-2.54}) \phantom{***} & \scorecell{57.91}{-1.13} ($\Delta$ \textcolor{myred}{-1.13}) \phantom{***} & \scorecell{56.21}{-1.13} ($\Delta$ \textcolor{myred}{-1.13}) \phantom{***} \\
 & \textit{Ru$\rightarrow$En} & \scorecell{60.45}{0.00} & \scorecell{59.60}{0.00} & \scorecell{60.17}{0.00} & \scorecell{61.02}{0.56} ($\Delta$ \textcolor{mygreen}{+0.56}) \phantom{***} & \scorecell{55.65}{-3.95} ($\Delta$ \textcolor{myred}{-3.95}) \phantom{***} & \scorecell{57.34}{-2.82} ($\Delta$ \textcolor{myred}{-2.82}) \phantom{***} \\
 & \textit{Ja$\rightarrow$En} & \scorecell{58.47}{0.00} & \scorecell{59.89}{0.00} & \scorecell{60.73}{0.00} & \scorecell{60.45}{1.98} ($\Delta$ \textcolor{mygreen}{+1.98}) \phantom{***} & \scorecell{59.04}{-0.85} ($\Delta$ \textcolor{myred}{-0.85}) \phantom{***} & \scorecell{60.73}{0.00} ($\Delta$ +0.00) \phantom{***} \\
\bottomrule
\end{tabular}
\caption{Accuracy (seed-mean of the final checkpoint, scored over tokens seen in L1 and/or L2 training data (OOV excluded)) under each vocabulary overlap conditions, for benchmarks CLAMS, MultiBLiMP. The $\Hier\to\Lin$ column appends $\Delta=(\Hier\to\Lin)-(\Lin\to\Lin)$ at the same overlap (\textcolor{mygreen}{green}=positive, \textcolor{myred}{red}=negative).}
\label{tab:result_overlap_clams_multiblimp_seen}
\end{table*}
\update{Table~\ref{tab:result_overlap_clams_multiblimp_seen} reports cross-lingual transfer under each vocabulary overlap condition; Figure~\ref{fig:clams_lang_delta} shows the corresponding $\Delta\mathrm{acc}$ by L1 and epoch.}
\begin{figure*}[t]
    \centering
      \includegraphics[width=0.9\linewidth]{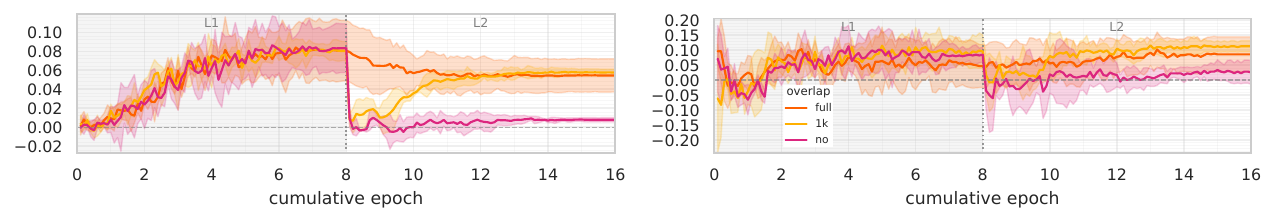}
      \caption{$\Delta\mathrm{acc}$ on CLAMS(Left) and MultiBLiMP(Right) for En$\to$En under extended training (8epoch in L1$\to$ 8epoch in L2). 
      Lines show the seed mean; shaded bands indicate$\pm 1$ SD across three random seeds.
      }
      \setlength{\belowcaptionskip}{-100pt}
      \label{fig:16epoch_en}
\end{figure*}

\begin{figure}[t]
\centering
\includegraphics[width=\linewidth]{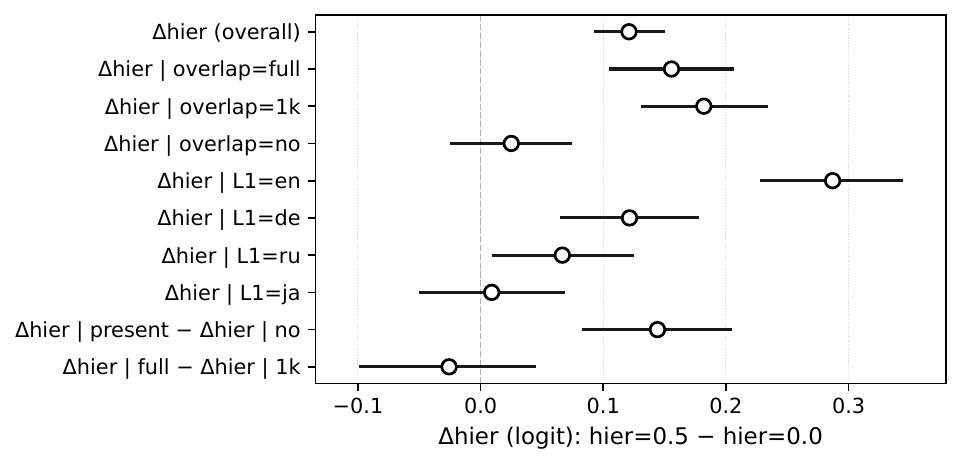}
\caption{\update{Principal transfer contrasts on CLAMS from the mixed-effects model. Points are posterior medians of $\Delta_{\mathrm{hier}}$ (log-odds) and error bars are 95\% CrIs; the dashed line marks zero. In the axis labels, hier$=$0.5 and hier$=$0.0 correspond to the \Hier$\to$\Lin{} and \Lin$\to$\Lin{} conditions, respectively.}}
\label{fig:glmm_forest_clams}
\end{figure}

Strikingly, whereas exposure to \update{\textbf{Disambiguating data}} in L1 yielded a reliable transfer under the \fulloverlap{} setting, the effect was almost entirely eliminated under the \nooverlap{} setting.
Even in the En$\rightarrow$En setting, the magnitude of $\Delta$ \update{is drastically reduced}.
Under \onekoverlap, which retains only a minimal lexical bridge, the En$\to$En case patterned with \nooverlap{} early in L2 training but recovered to a level on par with \fulloverlap{} by the late stages---a recovery that becomes especially clear when training is extended to 16 epochs (see Figures~\ref{fig:16epoch_en})\update{\footnote{The results for other L1 languages are shown in Figure~\ref{fig:16epoch_de}, ~\ref{fig:16epoch_ru}}.}
\update{On CLAMS, the mixed-effects analysis confirms this picture: the overall transfer contrast is credibly positive ($\Delta_{\mathrm{hier}}=0.121$ $[0.092, 0.150]$), and transfer is credibly stronger when vocabulary overlap is present than when it is absent ($\frac{\Delta_{\mathrm{hier}}^{(\textsc{Full})}+\Delta_{\mathrm{hier}}^{(\textsc{1K})}}{2}-\Delta_{\mathrm{hier}}^{(\textsc{No})}=0.144$ $[0.083, 0.205]$), whereas the contrast under \nooverlap{} is small and not credible ($\Delta_{\mathrm{hier}}^{(\textsc{No})}=0.025$ $[-0.025, 0.074]$; Figure~\ref{fig:glmm_forest_clams}).}
\update{At the L1 level on CLAMS, English and German settings show credible contrasts under vocabulary overlap, Japanese shows none, and under \nooverlap{} every credible interval includes zero (cell-level estimates in Table~\ref{tab:glmm_threeway_contrasts}, Appendix~\ref{app:glmm_details}).}
\update{On MultiBLiMP, the En$\to$En contrast is credible under \fulloverlap{} and \onekoverlap{} but not under \nooverlap{}, while all non-English credible intervals include zero (Table~\ref{tab:glmm_threeway_contrasts}); that is, the dependence of transfer on vocabulary overlap is not specific to the synthetic benchmark (CLAMS), but generally observed across several datasets.}

\section{Discussion}
\label{discussion}
Taken together, these observations indicate that cross-lingual transfer of hierarchical generalization \emph{does} take place, but is tightly constrained by the amount of shared vocabulary, and that even a minimal overlap is sufficient for the model to exploit as an anchor for transfer.

\section{Conclusion}
\label{sec:conclusion}
In this paper, we extended the study of hierarchical generalization beyond its conventional monolingual setting by casting L1-to-L2 continued pretraining as a cross-lingual poverty-of-stimulus design in which positive evidence for the hierarchical rule is supplied only in L1. 
Across four typologically diverse L1s, models exposed to \textbf{Disambiguating data} during L1 training exhibited above-chance hierarchical accuracy at L2 even when the L2 input contained no positive evidence and the Transformer was frozen, with the magnitude of transfer varying systematically across L1 languages.
Manipulating the L1--L2 shared-vocabulary inventory revealed that this transfer is tightly mediated by lexical overlap: making the two vocabularies fully disjoint nearly eliminated the effect, whereas a bridge of as few as $1{,}000$ shared tokens sufficed for the late-training effect to recover to fully shared levels. 
Taken together, these findings indicate that the inductive bias for hierarchical generalization acquired during pretraining is not strictly tied to a single language, yet its deployment in a new language is conditional on a lexical bridge between the two---a finding that bears directly on the design of multilingual tokenizers when cross-lingual reuse of syntactic abstractions is at stake.

\section*{Limitations}
\label{sec:limitation}

First, our experimental scope is limited in model size and language sample. All experiments use GPT-2-small; larger models and alternative architectures are not tested. 
Following \citet{arnett-etal-2025-acquisition}, we take the position that a phenomenon is best established at small scale before its scaling behaviour is examined. 
We further restrict L1 to four languages and fix L2 to English, chosen to span a gradient of word-order flexibility and attractor-licensing frequency; whether the findings generalize to typologically more diverse pairs, particularly outside Indo-European, remains untested. In particular, replicating the language gradient with L2 varied would be necessary to disentangle L1/L2 status from pair-specific lexical and structural proximity.

Second, the design controls that strengthen interpretability also constrain external validity. 
Freezing the Transformer during L2 training is what lets us isolate transfer of L1-acquired generalization from embedding-layer adaptation; our findings therefore do not extrapolate directly to standard multilingual pretraining or joint-training regimes, in which the Transformer is updated simultaneously. 
Whether the dependence on vocabulary overlap survives more flexible fine-tuning remains open. 
We also operationalize hierarchy through attractor-based subject-verb number agreement and clausal embedding depth, following the standard operationalization since \citet{linzen-etal-2016-assessing}; transfer on other hierarchical phenomena such as filler--gap dependencies or island constraints lies outside our present scope.

Third, synthetic and naturalistic evaluations diverge. 
The language gradient in \S\ref{sec:experiment_and_result} is \update{observed} on CLAMS, but on MultiBLiMP a \update{credible} effect appears only for English, with the sign of the German \update{point estimate} reversed. 
At least three accounts are compatible with this pattern: (i) the gradient is an artefact of statistical regularities specific to synthetic data; (ii) residual biases in naturalistic data---lexical frequency, collocational strength---mask the contribution of hierarchical generalization; or (iii) the smaller item counts of MultiBLiMP, together with three seeds, raise the minimal detectable effect beyond what our power supports. 
The fact that the English effect is reproduced on MultiBLiMP is compatible with (ii)+(iii), but our data do not adjudicate among the three. 
Disentangling them will require naturalistic benchmarks with larger item counts and replications with more seeds.

\section*{Ethical Statement}
We use AI services, including ChatGPT, Gemini, and Claude, for proofreading and writing assistance. Furthermore, we manually review the generated text to confirm that it contains no hallucinations, false statements, or reward-hacking code. Ultimately, the authors remain fully accountable for the accuracy and integrity of the entire text.
It is a study of language and language acquisition, and we did not intentionally generate or promote harmful content in our experiments.

\section*{Acknowledgements}
\update{We would like to express our gratitude for the anonymous reviewers who provided many insightful comments that have improved our paper. This work was supported by JSPS KAKENHI Grant Numbers 23K16938, 24K16077 and 26H02511.}

\bibliography{custom}

\appendix
\newpage
\begin{figure*}[t]
    \centering
      \includegraphics[width=0.8\linewidth]{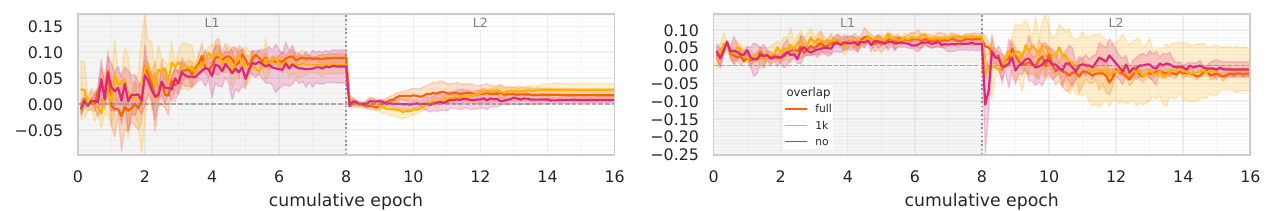}
      \caption{$\Delta\mathrm{acc}$ on CLAMS(Left) and MultiBLiMP(Right) for De$\to$En under extended training (8epoch in L1$\to$ 8epoch in L2). 
      Lines show the seed mean; shaded bands indicate$\pm 1$ SD across three random seeds.
      }
      \setlength{\belowcaptionskip}{-100pt}
      \label{fig:16epoch_de}
\end{figure*}
\begin{figure*}[t]
    \centering
      \includegraphics[width=0.8\linewidth]{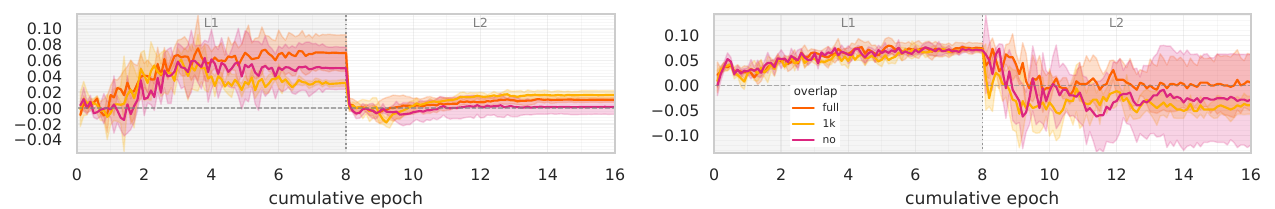}
      \caption{$\Delta\mathrm{acc}$ on CLAMS(Left) and MultiBLiMP(Right) for Ru$\to$En under extended training (8epoch in L1$\to$ 8epoch in L2). 
      Lines show the seed mean; shaded bands indicate$\pm 1$ SD across three random seeds.
      }
      \setlength{\belowcaptionskip}{-100pt}
      \label{fig:16epoch_ru}
\end{figure*}
\section{Supplementary Material}
\label{sec:appendix}
\subsection{Corpus Construction Details}
\label{app:corpus_construction_details}
\label{sec:data_construction_implementation}
We use Stanza~\cite{qi-etal-2020-stanza} both for token counting and positive evidence identification\footnote{\update{The performance of the pretrained Stanza models for all four languages is publicly reported on the official model-performance page. \url{https://stanfordnlp.github.io/stanza/performance.html}}}. 
The training data are drawn from Wikipedia/Wikimedia because (1) its domain closely matches that of the Universal Dependencies treebanks on which Stanza's parser was trained, yielding more reliable parses than developmentally plausible alternatives such as the child-directed speech in CHILDES, and (2) it is available at comparable scale across our four target languages, a prerequisite for the matched-size manipulation described in Section~\ref{subsec:data_construction}.

After parsing, we classify every sentence containing an
\texttt{nsubj} relation along two features that follow
the criteria in Section~\ref{sec:hierarchy_definition}.
\texttt{has\_attractors} (bool), coming from the implementation
of \citet{jumelet-etal-2026-multiblimp}, indicates whether any
noun appears linearly between the subject and the verb of the
\texttt{nsubj} relation. Following the
clausal embedding in \citet{blasi-etal-2019-distribution},
\texttt{embedding\_depth} (int) is
computed by traversing the head path from each word to the
sentence root and incrementing the count whenever the path
crosses a dependency relation that marks an embedded clause
(\texttt{acl}, \texttt{advcl}, \texttt{ccomp}, \texttt{xcomp},
\texttt{csubj}, and their subtypes); the sentence-level
\texttt{embedding\_depth} is the maximum of these per-word
depths.

\paragraph{Classification criteria.}
Sentences can be classified into one of the following two types.
\begin{description}
    \item[\update{\textbf{non-evidential sentence}}:] Sentences that do not contain attractors and have an embedding depth \update{of} 0.
    \item[\update{\textbf{positive-evidence sentence}}:] Sentences that contain attractors and have an embedding depth of at least 1.
\end{description}

\subsection{Training Corpus Statistics}
\label{app:corpus_stats}
Table~\ref{tab:datastats} reports the per-language sentence
counts and average sentence lengths for the \textbf{Ambiguous data} and
\textbf{Disambiguating data} partitions of the training corpora.

\begin{table*}[t]
\small
\centering
\begin{tabular}{l|cc|cc}
\toprule
Language
& \multicolumn{2}{c|}{\textbf{Ambiguous data}}
& \multicolumn{2}{c}{\textbf{Disambiguating data}} \\
\cmidrule(lr){2-3}
\cmidrule(lr){4-5}
& \#Sents & \#Words/Sent
& \#Sents & \#Words/Sent \\
\midrule
English (L1/L2) & 2.42/2.42M & 20.6/20.6 & 2.0M & 24.6 \\
German          & 2.7M       & 18.3      & 2.3M & 21.9 \\
Russian         & 2.8M       & 17.8      & 2.3M & 21.3 \\
Japanese        & 3.8M       & 13.1      & 2.8M & 17.9 \\
\bottomrule
\end{tabular}
\caption{Statistics of the training corpora. Each language group reports the number of sentences and the average sentence length for the \textbf{Ambiguous data} and \textbf{Disambiguating data} partitions. The English row additionally reports the L2 column since English serves as the L2 in all experiments.}
\label{tab:datastats}
\end{table*}
\begin{table*}[t]
\small
\centering
\begin{tabular}{l|rrrrr|rrrrr}
\toprule
Language
& \multicolumn{5}{c|}{\textbf{Ambiguous data}}
& \multicolumn{5}{c}{\textbf{Disambiguating data}} \\
\cmidrule(lr){2-6}
\cmidrule(lr){7-11}
& nsubj & Clausal & All & Depth & DLT Stor.
& nsubj & Clausal & All & Depth & DLT Stor. \\
\midrule
English  & 1.64 & 0.00 & 3.29 & 0.00 & 4.49 & 4.33 & 5.70  & 3.13 & 0.62 & 4.93 \\
German   & 1.38 & 0.00 & 3.21 & 0.00 & 4.24 & 5.44 & 10.64 & 3.47 & 0.48 & 5.38 \\
Russian  & 1.46 & 0.00 & 3.31 & 0.00 & 3.90 & 4.14 & 5.37  & 2.97 & 0.53 & 4.36 \\
Japanese & 2.30 & 0.00 & 2.87 & 0.00 & 2.55 & 7.06 & 5.27  & 2.88 & 0.66 & 2.96 \\
\bottomrule
\end{tabular}
\caption{\update{Dependency-based complexity statistics of the 50M-token L1 training corpora: nsubj = mean subject--verb linear distance over \texttt{nsubj} relations; Clausal = mean head--dependent distance over clausal relations (\texttt{acl}, \texttt{advcl}, \texttt{ccomp}, \texttt{xcomp}, \texttt{csubj}); All = the same over all relations except \texttt{punct}; Depth = mean per-sentence maximum clausal embedding depth; DLT Stor.\ = mean per-sentence maximum storage cost under the Dependency Locality Theory~\cite{gibson-2000-dependency}, i.e., the number of pending syntactic heads. Clausal and Depth are exactly $0$ in the \Lin$\to$\Lin{} corpora by construction. The English L2 corpus matches the English \Lin$\to$\Lin{} columns to two decimals.}} 
\label{tab:depstats}
\end{table*}
\subsection{Evaluation Benchmark Coverage}
\label{app:eval_coverage}
Table~\ref{tab:phenomena-hierarchical} lists the phenomena
included in the hierarchical partitions of CLAMS and
MultiBLiMP, together with the number of minimal-pair items
per phenomenon.

\begin{table}[ht]
\centering
\small
\begin{tabular}{lr}
\toprule
\textbf{Phenomenon (subcategory)} & \textbf{\# items} \\
\midrule
\multicolumn{2}{l}{\textbf{CLAMS} (4 phenomena)} \\
\midrule
\texttt{obj\_rel\_across\_anim} & 11,200 \\
\texttt{obj\_rel\_within\_anim} & 11,200 \\
\texttt{prep\_anim} & 16,800 \\
\texttt{subj\_rel} & 11,200 \\
\midrule
\textit{Total} & \update{\textit{50,400}} \\
\midrule
\multicolumn{2}{l}{\textbf{MultiBLiMP} (1 phenomenon)} \\
\midrule
\texttt{has\_attractors\_true} & 130 \\
\midrule
\textit{Total} & \textit{130} \\
\bottomrule
\end{tabular}
\caption{Phenomena included in the \emph{hierarchical} partition of CLAMS and MultiBLiMP, together with the number of minimal-pair items per phenomenon. These are the items aggregated into the accuracy reported in Section~\ref{sec:experiment_and_result}.}
\label{tab:phenomena-hierarchical}
\end{table}
\section{Details of Vocabulary Overlap}
\label{app:details_of_overlap}
\subsection{Implementation of Vocabulary Overlap.}
\label{app:vocabulary_overlap_implementation}
\begin{figure}[t]
    \centering
      \includegraphics[width=\linewidth]{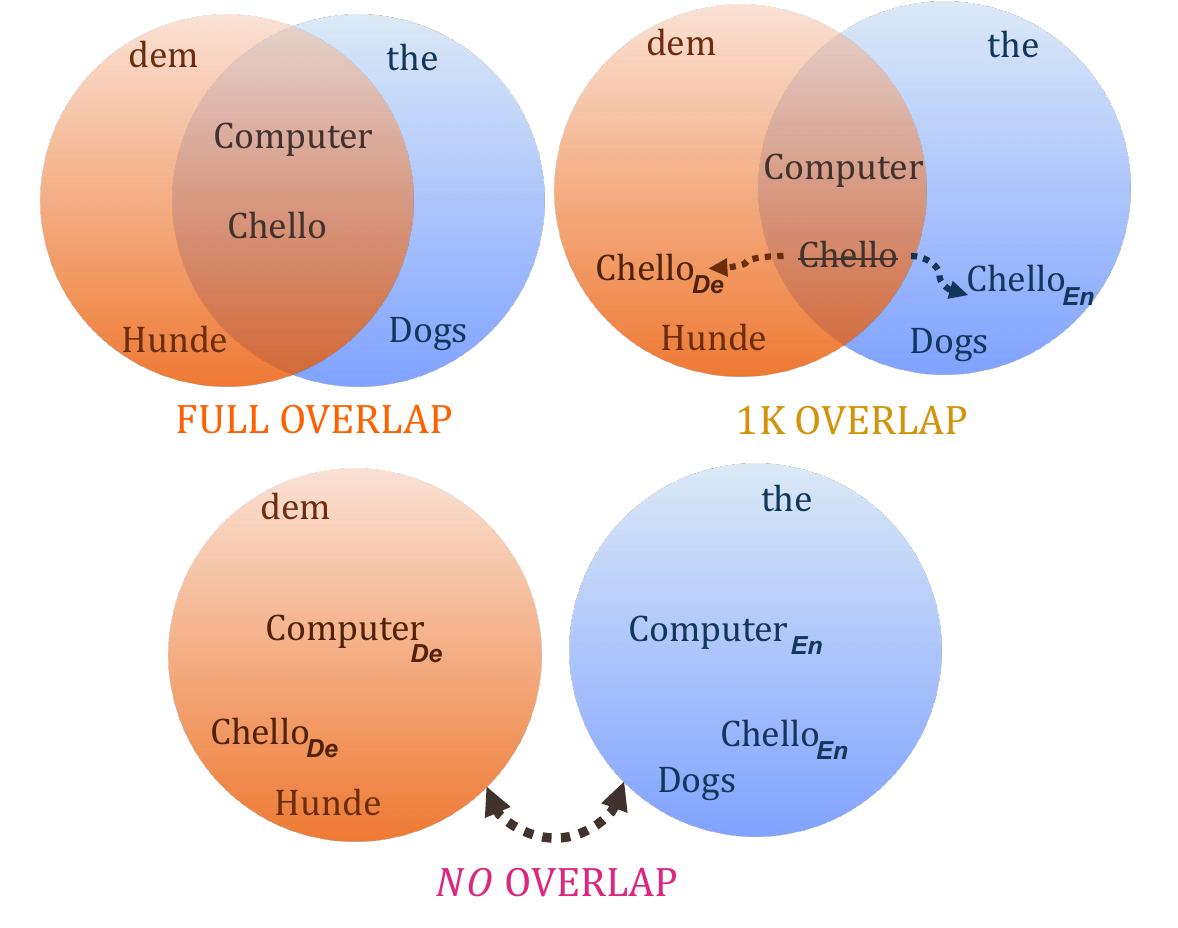}
      \caption{A visualization of the three overlap conditions. \fulloverlap, \nooverlap, \onekoverlap.
      }
      \setlength{\belowcaptionskip}{-100pt}
      \label{fig:overlap_implementation}
\end{figure}

Following the implementation of \citet{kallini-etal-2025-false},
we take the byte-level BPE tokenizer $\mathcal{T}$ trained on
the concatenation of L1 and L2 (Section~\ref{subsec:tokenizer}) as our
base and extract the native overlap
$O = V_{L_1} \cap V_{L_2}$, i.e.\ the set of tokens that
appear in both the L1 and L2 corpora. Each overlap condition
is fully specified by the subset $O' \subseteq O$ that is
retained as shared: tokens in $O \setminus O'$ are remapped
on the L2 side by an offset of $|O|$ so as to occupy a
disjoint ID range, while L2-specific (non-shared) tokens keep
their original IDs. Crucially, the offset amount is the
shared-vocabulary size $|O|$ rather than the base vocabulary
size $N$, ensuring that only the shared portion is displaced
into the disjoint region~\ref{fig:overlap_implementation}.
Thus, \fulloverlap
corresponds to $O' = O$ (no offset), \nooverlap to
$O' = \emptyset$ (all of $O$ offset), and \onekoverlap
to $|O'| = 1{,}000$, where $|O| - 1{,}000$ shared tokens are
offset and $1{,}000$ remain shared as a minimal lexical
anchor between the two languages. The resulting L1- and
L2-specific effective vocabulary sizes under each condition
are summarized in Table~\ref{tab:vocab-overlap}. Model
architecture, hyperparameters, and the training/evaluation
procedure are identical to those in
\S\ref{subsec:setup}.

\subsection{Vocabulary Overlap and Parameter Counts}
\label{app:vocab_overlap_details}
Table~\ref{tab:vocab-overlap} reports the L1- and L2-specific
effective vocabulary sizes and total parameter counts for
each (language pair, overlap setting) combination, separately
for the L1 \textbf{Ambiguous data} and \textbf{Disambiguating data}training corpora.

\begin{table*}[h]
\centering
\small
\resizebox{\textwidth}{!}{%
\begin{tabular}{ll|cc|cc|cc|cc|cc}
\toprule
& & \multicolumn{2}{c|}{$|V_1|$} & \multicolumn{2}{c|}{$|V_2|$} & \multicolumn{2}{c|}{$|O'|$} & \multicolumn{2}{c|}{\textbf{Vocabulary Size}} & \multicolumn{2}{c}{\textbf{Total Parameters (M)}} \\
\textbf{Language Pair} & \textbf{Setting} & Lin & Hier & Lin & Hier & Lin & Hier & Lin & Hier & Lin & Hier \\
\midrule
\multirow{3}{*}{En$\to$En} & \textcolor{fulloverlap}{Full Overlap} & \multirow{3}{*}{63,839} & \multirow{3}{*}{63,789} & \multirow{3}{*}{63,825} & \multirow{3}{*}{63,820} & 63,768 & 63,733 & 64,000 & 64,000 & 134.60 & 134.60 \\
 & \textcolor{1koverlap}{1k Overlap} &  &  &  &  & 1,000 & 1,000 & 126,768 & 126,733 & 182.81 & 182.78 \\
 & \textcolor{nooverlap}{No Overlap} &  &  &  &  & 0 & 0 & 127,768 & 127,733 & 183.58 & 183.55 \\
\midrule
\multirow{3}{*}{De$\to$En} & \textcolor{fulloverlap}{Full Overlap} & \multirow{3}{*}{62,676} & \multirow{3}{*}{62,238} & \multirow{3}{*}{56,707} & \multirow{3}{*}{55,048} & 55,473 & 53,374 & 64,000 & 64,000 & 134.60 & 134.60 \\
 & \textcolor{1koverlap}{1k Overlap} &  &  &  &  & 1,000 & 1,000 & 118,473 & 116,374 & 176.44 & 174.82 \\
 & \textcolor{nooverlap}{No Overlap} &  &  &  &  & 0 & 0 & 119,473 & 117,374 & 177.20 & 175.59 \\
\midrule
\multirow{3}{*}{Ru$\to$En} & \textcolor{fulloverlap}{Full Overlap} & \multirow{3}{*}{62,463} & \multirow{3}{*}{61,936} & \multirow{3}{*}{38,177} & \multirow{3}{*}{36,306} & 36,742 & 34,334 & 64,000 & 64,000 & 134.60 & 134.60 \\
 & \textcolor{1koverlap}{1k Overlap} &  &  &  &  & 1,000 & 1,000 & 99,742 & 97,334 & 162.05 & 160.20 \\
 & \textcolor{nooverlap}{No Overlap} &  &  &  &  & 0 & 0 & 100,742 & 98,334 & 162.82 & 160.97 \\
\midrule
\multirow{3}{*}{Ja$\to$En} & \textcolor{fulloverlap}{Full Overlap} & \multirow{3}{*}{63,015} & \multirow{3}{*}{62,336} & \multirow{3}{*}{37,487} & \multirow{3}{*}{37,862} & 36,665 & 36,354 & 64,000 & 64,000 & 134.60 & 134.60 \\
 & \textcolor{1koverlap}{1k Overlap} &  &  &  &  & 1,000 & 1,000 & 99,665 & 99,354 & 161.99 & 161.75 \\
 & \textcolor{nooverlap}{No Overlap} &  &  &  &  & 0 & 0 & 100,665 & 100,354 & 162.76 & 162.52 \\
\bottomrule
\end{tabular}%
}
\caption{Token statistics, vocabulary sizes, and parameter counts for each (language pair, overlap setting) combination, shown separately for the L1 \Lin\ (\textit{hier\_ratio}=0.0) and \Hier\ (\textit{hier\_ratio}=0.5) training corpora. $|V_1|$ and $|V_2|$ denote the L1- and L2-corpus vocabularies; $|O'|$ is the number of shared tokens retained after the disjoint strategy. Total parameters (in millions) are computed analytically from $\mathrm{vocab\_size}\times n_\text{embd}+n_\text{pos}\times n_\text{embd}+n_\text{layer}\times(12\,n_\text{embd}^2+13\,n_\text{embd})+2\,n_\text{embd}$.}
\label{tab:vocab-overlap}
\end{table*}

\subsection{Coverage of Overlap vocabulary in the Training Corpus}
Table \ref{tab:vocab-coverage} reports the coverage of shared and disjoint vocabulary under each condition of vocabulary overlap.
\begin{table*}[h]
\centering
\small
\resizebox{\textwidth}{!}{%
\begin{tabular}{ll|cc|cc|cc|cc|cc|cc}
\toprule
& & \multicolumn{6}{c|}{\textbf{L1 corpus tokens (\%)}} & \multicolumn{6}{c}{\textbf{L2 corpus tokens (\%)}} \\
& & \multicolumn{2}{c|}{Shared} & \multicolumn{2}{c|}{Disjointed} & \multicolumn{2}{c|}{L1-unique} & \multicolumn{2}{c|}{Shared} & \multicolumn{2}{c|}{Disjointed} & \multicolumn{2}{c}{L2-unique} \\
\textbf{Language Pair} & \textbf{Setting} & Lin & Hier & Lin & Hier & Lin & Hier & Lin & Hier & Lin & Hier & Lin & Hier \\
\midrule
\multirow{3}{*}{En$\to$En} & \textcolor{fulloverlap}{Full Overlap} & 100.00 & 100.00 & 0.00 & 0.00 & \multirow{3}{*}{0.00} & \multirow{3}{*}{0.00} & 100.00 & 99.99 & 0.00 & 0.00 & \multirow{3}{*}{0.00} & \multirow{3}{*}{0.01} \\
 & \textcolor{1koverlap}{1k Overlap} & 60.06 & 59.26 & 39.94 & 40.74 &  &  & 60.05 & 59.98 & 39.94 & 40.01 &  &  \\
 & \textcolor{nooverlap}{No Overlap} & 0.00 & 0.00 & 100.00 & 100.00 &  &  & 0.00 & 0.00 & 100.00 & 99.99 &  &  \\
\midrule
\multirow{3}{*}{De$\to$En} & \textcolor{fulloverlap}{Full Overlap} & 94.32 & 92.56 & 0.00 & 0.00 & \multirow{3}{*}{5.68} & \multirow{3}{*}{7.44} & 99.04 & 98.70 & 0.00 & 0.00 & \multirow{3}{*}{0.96} & \multirow{3}{*}{1.30} \\
 & \textcolor{1koverlap}{1k Overlap} & 43.44 & 39.32 & 50.87 & 53.24 &  &  & 50.94 & 50.96 & 48.10 & 47.74 &  &  \\
 & \textcolor{nooverlap}{No Overlap} & 0.00 & 0.00 & 94.32 & 92.56 &  &  & 0.00 & 0.00 & 99.04 & 98.70 &  &  \\
\midrule
\multirow{3}{*}{Ru$\to$En} & \textcolor{fulloverlap}{Full Overlap} & 71.12 & 66.70 & 0.00 & 0.00 & \multirow{3}{*}{28.88} & \multirow{3}{*}{33.30} & 98.75 & 98.22 & 0.00 & 0.00 & \multirow{3}{*}{1.25} & \multirow{3}{*}{1.78} \\
 & \textcolor{1koverlap}{1k Overlap} & 27.80 & 22.60 & 43.32 & 44.10 &  &  & 49.74 & 48.96 & 49.01 & 49.26 &  &  \\
 & \textcolor{nooverlap}{No Overlap} & 0.00 & 0.00 & 71.12 & 66.70 &  &  & 0.00 & 0.00 & 98.75 & 98.22 &  &  \\
\midrule
\multirow{3}{*}{Ja$\to$En} & \textcolor{fulloverlap}{Full Overlap} & 76.42 & 73.27 & 0.00 & 0.00 & \multirow{3}{*}{23.58} & \multirow{3}{*}{26.73} & 99.18 & 98.62 & 0.00 & 0.00 & \multirow{3}{*}{0.82} & \multirow{3}{*}{1.38} \\
 & \textcolor{1koverlap}{1k Overlap} & 17.74 & 12.93 & 58.68 & 60.34 &  &  & 49.71 & 48.59 & 49.47 & 50.03 &  &  \\
 & \textcolor{nooverlap}{No Overlap} & 0.00 & 0.00 & 76.42 & 73.27 &  &  & 0.00 & 0.00 & 99.18 & 98.62 &  &  \\
\bottomrule
\end{tabular}%
}
\caption{Per-corpus token coverage (\%) of the four dissociation categories. Cells are the share of the 50M-word L1 (resp.\ L2) training corpus whose tokens fall into each category, computed by tokenizing each corpus with its (language pair, hier\_ratio)-specific joint Byte-Level BPE. Categories: \emph{Shared} are intersection tokens whose vocabulary IDs are kept identical across L1 and L2; \emph{Disjointed} are intersection tokens whose L2-side IDs are shifted into the $[V,V{+}D)$ range, giving them separate embeddings; \emph{L1- / L2-unique} are tokens that appear only in the respective corpus. L1-unique on the L1 side (and L2-unique on the L2 side) are constant across overlap settings within a (pair, hier) cell and shown via \texttt{multirow}; the complementary cells (L2-unique on the L1 side and vice versa) are zero by definition and omitted. Within each (pair, hier) the three pct values (Shared, Disjointed, language-specific unique) sum to $\approx 100\%$ on each side.}
\label{tab:vocab-coverage}
\end{table*}

\section{Hyperparameters}
\label{app:hyperparameters}
Table~\ref{tab:hyperparameter} reports the architectural and
optimization hyperparameters used in all experiments. The
same configuration is used across L1 and L2 training, the
only difference being that L2 training freezes the
Transformer layers, as described in
Section~\ref{subsec:setup}.

\begin{table}[h]
\small 
    \centering
    \begin{tabular}{l|c}
    \toprule
     Hyperparameter    & Setting \\
    \midrule 
    \update{Training-corpus size per stage} & 50M tokens \\
    Number of layers ($n_\text{layer}$) & 12 \\
Number of attention heads ($n_\text{head}$) & 12 \\
Hidden size ($n_\text{embd}$) & 768 \\
     learning rate    &  5e-4\\
     Per-device batch size & 32 \\
Gradient accumulation steps & 2 \\
     context length & 512\\
     warmup steps & 1,500\\
     \update{Epochs per stage} & 8 \\
     dropout & 0.1\\
     weight decay & 0.1 \\
     learning rate scheduler & cosine \\
     \midrule
\multicolumn{2}{l}{\text{Evaluation}} \\
\midrule
Checkpoint save interval & every 1 epoch \\
Validation interval & every 0.05 epoch \\
Benchmark evaluation interval & every 0.1 epoch \\
     
     \bottomrule
    \end{tabular}
    \caption{Hyperparameter settings used for the experiments.}
    \label{tab:hyperparameter}
\end{table}

\begin{figure*}[t]
    \centering
      \includegraphics[width=\linewidth]{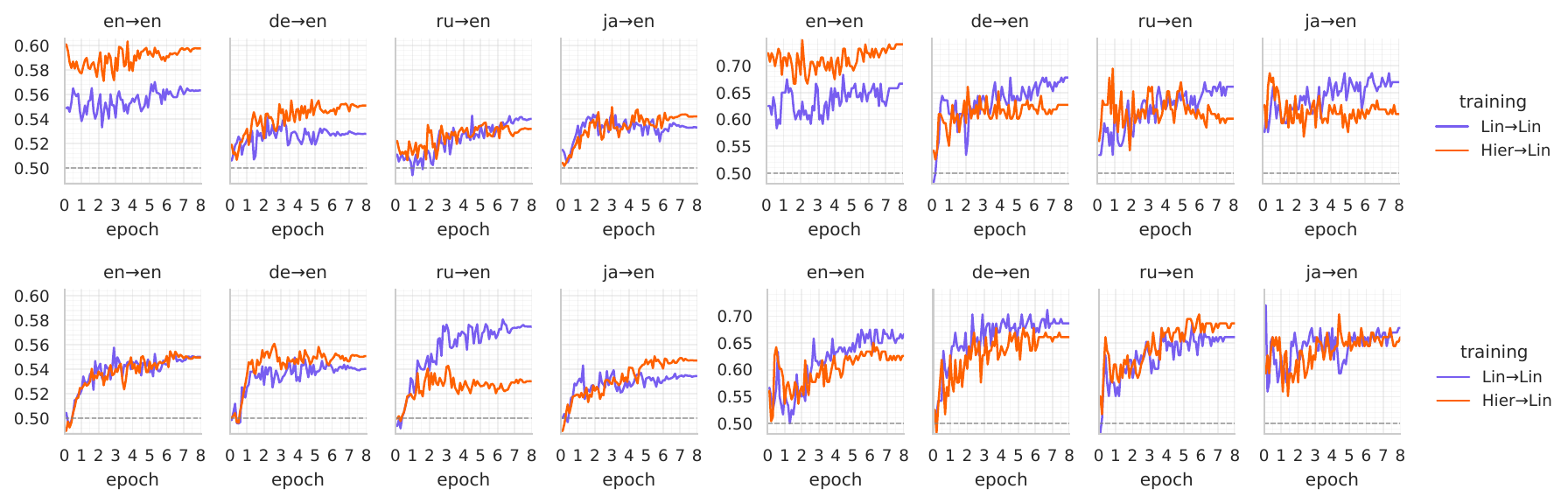}
      \caption{\update{Results of the unfrozen-Transformer ablation under \fulloverlap{} (top) and \nooverlap{} (bottom); all Transformer layers are updated during L2 training, while the remaining settings match the main experiment.}}
      \setlength{\belowcaptionskip}{-100pt}
      \label{fig:unfreezing}
\end{figure*}

\update{
\section{Details of the Statistical Analysis}
\label{app:glmm_details}We fit a separate Bayesian Bernoulli-logit mixed-effects model for each benchmark using all 72 independently trained runs ($2$ L1-evidence conditions $\times$ $3$ overlap conditions $\times$ $4$ L1 languages $\times$ $3$ seeds).
For response $y_n$ in cell $c[n]$, item $i[n]$, and run $r[n]$, the model is $y_n\sim\mathrm{Bernoulli}(\operatorname{logit}^{-1}(\mu_{c[n]}+a_{i[n]}+b_{r[n]}))$, where $c$ indexes the L1-evidence $\times$ overlap $\times$ L1-language cells.
We use cell-means parameterization of the fully interacted model in Section~\ref{sec:analysis}, with one mean for each of the 24 L1-evidence $\times$ overlap $\times$ L1-language combinations; $\Delta_{\mathrm{hier}}$ is the \Hier$\to$\Lin{} minus \Lin$\to$\Lin{} contrast between the corresponding posterior cell means.
The item and run intercepts account for repeated evaluation of the same item across conditions and dependence among items evaluated with the same trained model, respectively, where a run is one L1-evidence $\times$ overlap $\times$ L1-language $\times$ seed realization.
We use weakly informative priors, with $\mu_c\sim\mathrm{Normal}(0,2.5)$, $a_i\sim\mathrm{Normal}(0,\sigma_{\mathrm{item}})$, $b_r\sim\mathrm{Normal}(0,\sigma_{\mathrm{run}})$, $\sigma_{\mathrm{item}}\sim\mathrm{HalfNormal}(1.0)$, and $\sigma_{\mathrm{run}}\sim\mathrm{HalfNormal}(0.5)$.
Sampling used four chains with 4{,}000 posterior draws each; there were no divergent transitions and all $\hat{R}\leq 1.01$.
}
\begin{table*}[t]
    \centering
    \small
    \begin{subtable}[t]{0.48\textwidth}
    \centering
    \begin{adjustbox}{max width=\linewidth}
    \begin{tabular}{@{}llrr@{}}
    \toprule
    L1 & Overlap & Estimate & 95\% CrI \\
    \midrule
    \multirow{3}{*}{En} & \textsc{Full} & $\mathbf{0.390}$ & $\mathbf{[0.289,\ 0.489]}$ \\
     & \textsc{1k} & $\mathbf{0.414}$ & $\mathbf{[0.311,\ 0.520]}$ \\
     & \textsc{No} & $0.055$ & $[-0.042,\ 0.156]$ \\
    \addlinespace
    \multirow{3}{*}{De} & \textsc{Full} & $\mathbf{0.119}$ & $\mathbf{[0.017,\ 0.221]}$ \\
     & \textsc{1k} & $\mathbf{0.191}$ & $\mathbf{[0.089,\ 0.293]}$ \\
     & \textsc{No} & $0.054$ & $[-0.045,\ 0.152]$ \\
    \addlinespace
    \multirow{3}{*}{Ru} & \textsc{Full} & $0.076$ & $[-0.024,\ 0.174]$ \\
     & \textsc{1k} & $\mathbf{0.115}$ & $\mathbf{[0.013,\ 0.216]}$ \\
     & \textsc{No} & $0.010$ & $[-0.089,\ 0.108]$ \\
    \addlinespace
    \multirow{3}{*}{Ja} & \textsc{Full} & $0.039$ & $[-0.065,\ 0.141]$ \\
     & \textsc{1k} & $0.007$ & $[-0.092,\ 0.110]$ \\
     & \textsc{No} & $-0.018$ & $[-0.122,\ 0.085]$ \\
    \addlinespace
    \midrule
    \multicolumn{4}{@{}l}{\textit{Overlap-presence interaction by L1}} \\
    En & Present $-$ No & $\mathbf{0.346}$ & $\mathbf{[0.220,\ 0.470]}$ \\
    De & Present $-$ No & $0.102$ & $[-0.021,\ 0.222]$ \\
    Ru & Present $-$ No & $0.085$ & $[-0.036,\ 0.208]$ \\
    Ja & Present $-$ No & $0.042$ & $[-0.083,\ 0.168]$ \\
    \bottomrule
    \end{tabular}
    \end{adjustbox}
    \caption{CLAMS}
    \end{subtable}
    \hfill
    \begin{subtable}[t]{0.48\textwidth}
    \centering
    \begin{adjustbox}{max width=\linewidth}
    \begin{tabular}{@{}llrr@{}}
    \toprule
    L1 & Overlap & Estimate & 95\% CrI \\
    \midrule
    \multirow{3}{*}{En} & \textsc{Full} & $\mathbf{0.627}$ & $\mathbf{[0.217,\ 1.049]}$ \\
     & \textsc{1k} & $\mathbf{0.840}$ & $\mathbf{[0.425,\ 1.259]}$ \\
     & \textsc{No} & $0.188$ & $[-0.217,\ 0.596]$ \\
    \addlinespace
    \multirow{3}{*}{De} & \textsc{Full} & $-0.160$ & $[-0.562,\ 0.249]$ \\
     & \textsc{1k} & $-0.081$ & $[-0.486,\ 0.328]$ \\
     & \textsc{No} & $-0.078$ & $[-0.481,\ 0.334]$ \\
    \addlinespace
    \multirow{3}{*}{Ru} & \textsc{Full} & $0.018$ & $[-0.388,\ 0.429]$ \\
     & \textsc{1k} & $-0.279$ & $[-0.690,\ 0.131]$ \\
     & \textsc{No} & $-0.199$ & $[-0.602,\ 0.210]$ \\
    \addlinespace
    \multirow{3}{*}{Ja} & \textsc{Full} & $0.141$ & $[-0.268,\ 0.538]$ \\
     & \textsc{1k} & $-0.060$ & $[-0.468,\ 0.348]$ \\
     & \textsc{No} & $0.039$ & $[-0.366,\ 0.444]$ \\
    \addlinespace
    \midrule
    \multicolumn{4}{@{}l}{\textit{Overlap-presence interaction by L1}} \\
    En & Present $-$ No & $\mathbf{0.545}$ & $\mathbf{[0.047,\ 1.048]}$ \\
    De & Present $-$ No & $-0.045$ & $[-0.539,\ 0.449]$ \\
    Ru & Present $-$ No & $0.070$ & $[-0.425,\ 0.562]$ \\
    Ja & Present $-$ No & $0.003$ & $[-0.505,\ 0.505]$ \\
    \bottomrule
    \end{tabular}
    \end{adjustbox}
    \caption{MultiBLiMP}
    \end{subtable}
    \caption{
    \update{Model-estimated transfer contrasts and planned overlap-presence difference-in-differences by L1.
    Rows report the model estimate of $\Delta_{\mathrm{hier}}$ within each vocabulary-overlap $\times$ L1 combination; ``Present $-$ No'' denotes $(\Delta_{\textsc{Full}}+\Delta_{\textsc{1K}})/2-\Delta_{\textsc{No}}$.
    Entries are estimates (posterior medians) with equal-tailed 95\% CrIs on the log-odds scale, and positive estimates indicate a higher probability of a correct response under \Hier$\to$\Lin{} than under \Lin$\to$\Lin{}.
    \textbf{Bold} entries have CrIs that lie entirely above zero and are therefore described as credibly positive.}
    }
    \label{tab:glmm_threeway_contrasts}
\end{table*}
\update{
\section{Unfreezing Experiment}
\label{app:unfreezing}
To test whether freezing the Transformer during L2 training drives the vocabulary-overlap effect, we keep the remaining training and evaluation settings identical to the main experiment and change only whether the Transformer layers are updated during L2 training (Figure~\ref{fig:unfreezing}).
The ablation compares the \fulloverlap{} and \nooverlap{} conditions.}

\section{Computational Resources}
Across all experiments, we used a total of approximately 1116 GPU hours. Training was conducted on NVIDIA A100 80GB and NVIDIA A100 40GB GPUs.
\end{document}